\documentclass[pdflatex,sn-mathphys-num]{sn-jnl}

\usepackage{graphicx}%
\usepackage{multirow}%
\usepackage{amsmath,amssymb,amsfonts}%
\usepackage{amsthm}%
\usepackage{mathrsfs}%
\usepackage[title]{appendix}%
\usepackage{xcolor}%
\usepackage{textcomp}%
\usepackage{manyfoot}%
\usepackage{booktabs}%
\usepackage{algorithm}%
\usepackage{algorithmicx}%
\usepackage{algpseudocode}%
\usepackage{listings}%

\usepackage{subcaption}

\theoremstyle{thmstyleone}%
\theoremstyle{thmstyletwo}%

\theoremstyle{thmstylethree}%

\begin{document}

\title[Article Title]{A Novel BERT-based Classifier to Detect Political Leaning of YouTube Videos based on their Titles}


\author[1]{\fnm{Nouar} \sur{AlDahoul}}\email{naa9497@nyu.edu}

\author[1]{\fnm{Talal} \sur{Rahwan}}\email{talal.rahwan@nyu.edu}

\author*[1]{\fnm{Yasir} \sur{Zaki}}\email{yasir.zaki@nyu.edu}

\affil*[1]{\orgdiv{Computer Science}, \orgname{New York University Abu Dhabi}, \orgaddress{\country{UAE}}}


\abstract{A quarter of US adults regularly get their news from YouTube. Yet, despite the massive political content available on the platform, to date no classifier has been proposed to identify the political leaning of YouTube videos. To fill this gap, we propose a novel classifier based on Bert---a language model from Google---to classify YouTube videos merely based on their titles into six categories, namely: Far Left, Left, Center, Anti-Woke, Right, and Far Right. We used a public dataset of 10 million YouTube video titles (under various categories) to train and validate the proposed classifier. We compare the classifier against several alternatives that we trained on the same dataset, revealing that our classifier achieves the highest accuracy (75\%) and the highest F1 score (77\%). To further validate the classification performance, we collect videos from YouTube channels of numerous prominent news agencies, such as Fox News and New York Times, which have widely known political leanings, and apply our classifier to their video titles. For the vast majority of cases, the predicted political leaning matches that of the news agency. }


\maketitle

\section{Introduction}\label{sec1}
The widespread use of the World Wide Web has led to a substantial increase in the number of adults who consume at least some of their news online, reaching as high as 89\% among adults in the United States~\cite{PewResearch}. YouTube, one of the most popular websites on the World Wide Web, is rapidly growing its content, with more than 500 hours of video being uploaded every minute, amounting to a total of about 30,000 hours of new content every hour~\cite{17}. There are currently more than two billion users using the platform, and \textit{YouTube Shorts} alone have received 70 billion views to date, according to~\cite{1}. Politics is among the many topics covered by the platform. A quarter of adults in the US regularly get their news from YouTube, making it the second most popular online news source worldwide~\cite{2,3}.

Several studies have demonstrated the political leaning and bias in media, particularly news articles~\cite{8,9,10,11,12}. These studies proposed classifiers to predict the bias using textual data extracted from headlines or contents. In the context of YouTube, there have been numerous solutions aimed at categorizing videos into various classes~\cite{18,19,20,21}. These solutions focused on classifying news documents or video titles using conventional machine learning algorithms. However, none of them have used transformer-based embedding models, which have proven to be the state of the art in multiple domains~\cite{22,23,24,25}. Moreover, no classifier has been proposed to date to identify the political leaning of YouTube videos based solely on their title.

The capability of embedding models to learn left-to-right and right-to-left contexts and produce a meaningful representation has been a challenge for a long time. Google's BERT is a language model that addresses this challenge by learning a bidirectional representation. Having an effective representation or embedding of text is a key factor in building a highly accurate text classifier. BERT has shown superior performance as an embedding model for various classification purposes~\cite{22,23,24,25}. Language models require a large dataset to train on in order to avoid the problem of overfitting. Fortunately, in our context of classifying the political leaning of YouTube videos, a large dataset already exists, consisting of 11.5 million videos labelled based on their political leaning~\cite{5,6,7}.

Prior works utilized traditional ML algorithms for embedding, such as term frequency-inverse document frequency (TF-IDF)~\cite{gu2021prediction}, word2Vec~\cite{9396875} and the Glove model~\cite{xiao2023detecting}. These previous models have limitations in finding informative word representations from context~\cite{di2021considerations}; as such, this has an impact on the accuracy of the classification task. This problem can be found in several works, such as fake news detection~\cite{essa2023fake}, text sentiment analysis~\cite{shen2021comparison}, and topic classification~\cite{wang2020comparative}.

\begin{figure}[htbp]
    \centering
    \includegraphics[width=.9\linewidth]{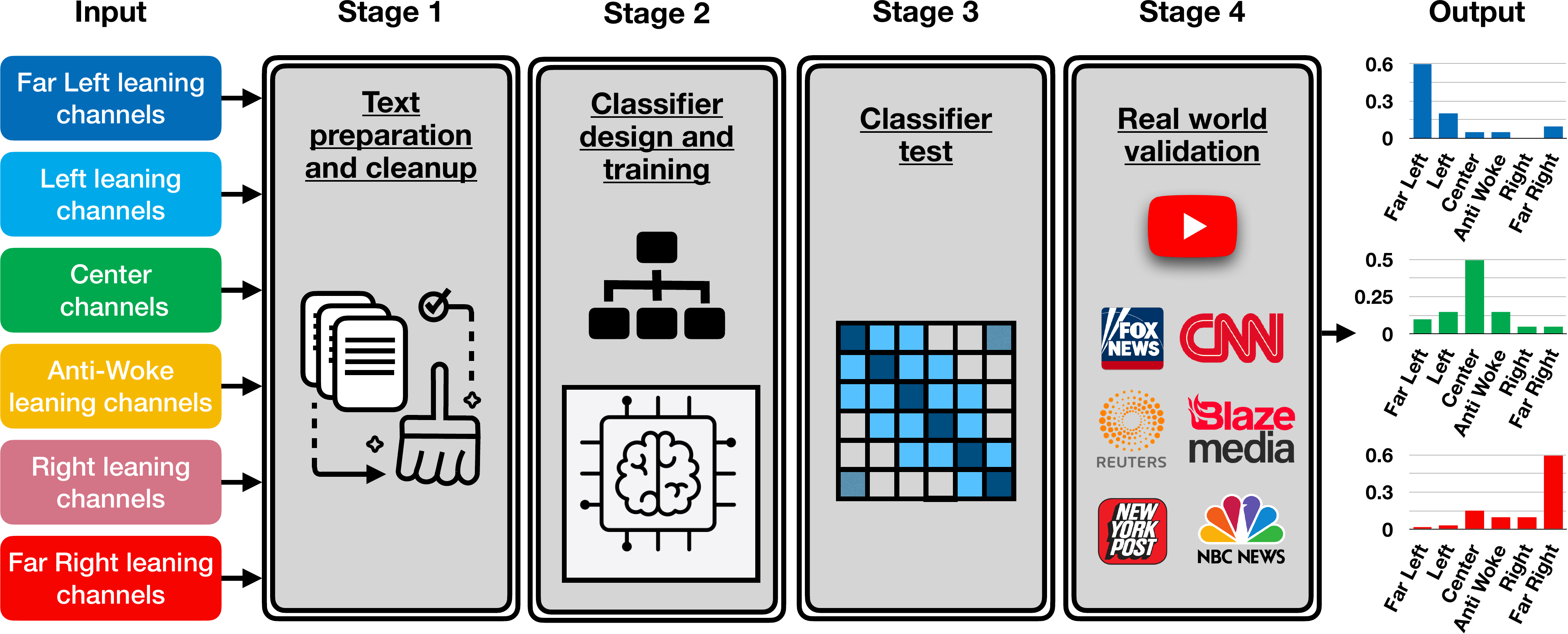}
    \caption{\textbf{Experimental setup.} An illustration of the different stages undertaken during our experiment. In Stage~1, the labelled video titles are prepared and cleaned. In Stage~2, the classification model is designed, trained, and validated utilizing the video title dataset. In Stage~3, the model is tested using a separated set of video titles to evaluate its performance. Finally, in Stage~4, video titles collected from 15 YouTube channels are used for model validation.}
    \label{fig:setup}
\end{figure}

Furthermore, many of the previous works focused solely on the detection of political leaning in newspaper articles~\cite{8}, tweets~\cite{jiang2023retweet}, and Facebook~\cite{nyhan2023like}. However, there are no prior studies that used the titles of YouTube videos to detect the political leaning of such videos. Here, to fill the above gap in the literature, we explore three pre-trained text classifiers, namely Word2Vec~\cite{13}, Global Vectors for Word Representation (GloVe)~\cite{14}, and Bidirectional Encoder Representations from Transformers (BERT)~\cite{15}. We fine-tune these classifiers using the aforementioned dataset, where videos are pre-labelled into six classes, namely Far Left, Left, Center, Anti-Woke, Right, and Far Right. To validate our proposed approach, we analyze the video content of 15 channels that have widely known political leanings. These are divided into five channels with Left leaning, five channels with Right leaning, and five channels with Center leaning. Thousands of videos from each channel have been collected to extract the titles along with the dates on which the videos were uploaded. Figure~\ref{fig:setup} illustrates the pipeline used in our experiments. The outcome of this study is a novel video title classifier that generates a political leaning distribution for each channel.

\newpage 
The main contributions of this work are summarized as follows:
\begin{itemize}
    \item This is the first work that targets the YouTube platform, and in particular the video titles, to detect political leaning by fine-tuning three pre-trained text classifiers on a large-scale video title dataset.
    \item Our proposed fine-tuned BERT classifier was validated with thousands of videos collected from 15 YouTube channels of prominent news agencies.
\end{itemize}

This paper is organized as follows: Section ``Materials and Methods'' describes the dataset and discusses numerous text classification models, such as Word2Vec, GloVe, and BERT. Section ``Experiments and results'' describes the experiments conducted to evaluate various text classifiers. Finally, Section ``Conclusion and Future Work'' summarizes the work and discusses potential future directions.

\section*{Related Work}

Several research articles have studied the political leaning in media for various applications and use cases. One such application is algorithmic recommendations~\cite{4}. This study examined YouTube’s recommendation algorithm in the context of U.S.\ politics, to determine whether the algorithm is neutral, or whether it leans in a certain political direction. The authors found evidence that the recommendation algorithm is left-leaning, as it pulls users away from Far Right content stronger than from Far Left content. Another application in which the examination of political leaning can be helpful is the study of radical content consumption~\cite{5}. Here, the authors showed that the trends in video-based political news consumption are determined by various factors, the most important of which is individual preferences.

Perhaps the application that is most relevant to the context of our study is the prediction of political leaning in videos, which has been explored by numerous articles~\cite{5,8,9,10,11,12}. Specifically, in~\cite{5}, a binary random forest classifier composed of 96 predictors was trained. To identify the political leaning of any given video, the authors utilize a feature engineering method by analyzing the web partisan score of news domains viewed by users before and after the video in question, as well as the political leaning of all videos watched within the same session. The authors also rely on user-level features, such as the individual’s monthly consumption and web categories. In~\cite{8}, the authors proposed a generalized SVD-modeling of phrase statistics to infer a leaning conditional probability distribution in a given newspaper article. In~\cite{9}, Kulkarni et al.\ explore the possibility of using an article’s title and link structure to predict any biases therein. The authors capture cues from both textual content and the network structure of news articles using a novel attention-based multi-view model. In~\cite{10}, Li and Goldwasser demonstrate how social contents could be utilized to improve bias prediction by using graph convolutional networks to encode a social network graph. The study of political bias has been extended to other languages such as German and Indian~\cite{11,12}. More specifically, a dataset of German news articles labelled by a fine-grained set of labels was utilized for political bias classification~\cite{11}. The authors explored various feature extraction models, including bag-of-words, term-frequency times, inverse-document-frequency, and BERT, along with various classifiers, including logistic regression, naive Bayes, and random forest. Gangula et al.~\cite{12} analyzed news articles in the Indian language Telugu to detect political bias using 1,329 headlines of articles. The authors compared several models, such as Convolutional Neural Network (CNN), Long Short-term Memory (LSTM), and attention network, and found the latter outperformed the former models.

Recently, the BERT model has been used in several studies for the purpose of detecting political leaning. For example, the authors of~\cite{jiang2023retweet} used BERT to study the political discourse on Twitter. The authors utilized the ``RetweetBERT'' model to estimate the political leanings of Twitter users using their profile descriptions. Similarly, the authors of~\cite{nyhan2023like} estimated the political leaning of US adult Facebook users. Specifically, they utilized ``DistillBERT'', which is an externally trained classifier on Facebook content using text-based features and text extracted from images using Optical Character Recognition (OCR) techniques, to generate predictions for Facebook posts that were created, seen, or engaged. The classifier produced predictions at the user level ranging from 0 (left-leaning) to 1 (right-leaning).

Despite the usage of BERT in these previous studies as a classifier to detect political leaning, many of them focused only on a few social media platforms, such as Facebook and Twitter. There are no prior studies that targeted YouTube videos to automatically detect their political leaning using only their titles, which is the aim of our work.
 
\section*{Materials and Methods}
\subsection*{Data Overview}

The classification of the political leaning of YouTube videos has been examined in two studies, each using a different categorization of videos~\cite{6,7}. To unify the categories used in this context, Hosseinmardi et al.~\cite{5} proposed a dataset of 11.5 million YouTube videos that were collected in 2016–2019 and labelled into six political categories, namely: Far Left, Left, Center, Anti-Woke, Right, Far Right. The vast majority of videos in this dataset are primarily concerned with the U.S.\ political zeitgeist. It should be noted that the videos are classified based on the political leaning of the channels they fall under, rather than the video itself. For instance, given a channel that is categorized as Left, all videos therein are also categorized as Left. While this approach has the advantage of being scalable, it admittedly could assign inaccurate labels to some videos whose leanings may differ from those of the channel they fall under.

\begin{figure}[htbp]
    \centering
    \includegraphics[width=0.8\linewidth]{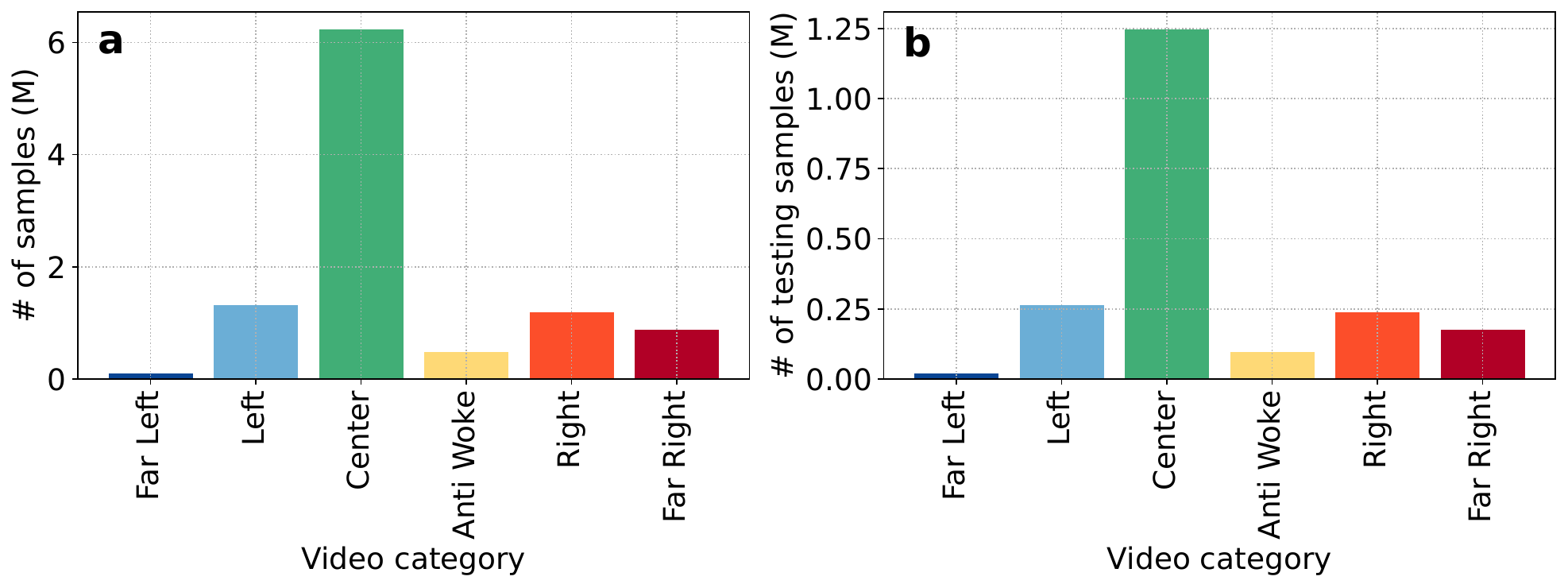}
    \caption{\textbf{Distribution of categories in our dataset.} The left plot depicts the distribution of category in the entire dataset, while the right plot depicts the distribution in the testing dataset.}
    \label{fig:data_distribution}
\end{figure}

In our experiment, we use the dataset of Hosseinmardi et al.~\cite{5}. We retrieved the titles of the videos therein, and cleaned them to avoid duplicates and missing values, resulting in a dataset consisting of 10,216,502 video titles. We utilized these text titles to train and evaluate three text classifiers; see Methods for more details. Figure~\ref{fig:data_distribution}a depicts the distribution of the six political categories in our dataset, showing that the dataset is imbalanced, with the majority of videos falling under the Center category. Figure~\ref{fig:data_distribution}b shows that the testing dataset exhibits a similar imbalance. Thus, to obtain high prediction accuracy, it is essential for the training stage to take into consideration this imbalance.

The 10,216,502 video titles in our dataset were split into three disjoint sets: (i) a training set consisting of 6,538,557 titles used to train the text classifier on video titles; (ii) a validation set consisting of 1,634,642 titles used to validate the classifier, optimize the architecture, and fine-tune the hyperparameters; and (iii) a testing set consisting of 2,043,303 titles used to evaluate the classifier prediction capability.

\subsection*{Methods}

This section describes the algorithm, architecture, and hyperparameters used in our experiments. It also describes the three embedding models that we use, namely Word2Vec~\cite{13}, GloVe~\cite{14}, and BERT~\cite{15}. Each of these models has its own algorithm and architecture. We conducted several experiments to determine the optimal architecture of each model, i.e., the one that yields the highest accuracy based on the validation data. To build video title text classification models, we added other layers, such as convolutional 1-D, LSTM, and dense layers, to the embedding models. Furthermore, we utilized a weighted loss function to assign more weights to the classes that have minority samples compared to those assigned to majority samples. This technique is commonly used when dealing with imbalanced datasets~\cite{fernando2021dynamically}.

\subsection*{Word2Vec}
Word2vec is a Natural Language Processing (NLP) technique that utilizes a shallow, two-layer neural network that is trained to reconstruct the linguistic contexts of words. Word2Vec usually learns word representations by representing each word in a large corpus of text as a vector called an embedding vector. By using this technique, we can capture the semantic and syntactic qualities of words by calculating the cosine similarity between the words represented by embedding vectors~\cite{13}.

In our experiment, we use a Word2Vec embedding model trained on a Google News dataset with a corpus of six billion tokens and a vocabulary size of one million, consisting of the most frequent words~\cite{13}. The model was fine-tuned on our video title dataset, with 700,000 vocabularies and a maximum sentence length of 100. Each word is represented by 300 dimensions. A sequence of layers was used, including a convolutional 1-dimensional layer, a batch normalization layer, and a max pooling layer, followed by two dense layers. The last dense layer produced six probabilities for the six political leaning categories, i.e., Far Left, Left, Center, Anti-Woke, Right, and Far Right. This architecture is the one that yielded the highest validation accuracy upon optimizing the hyperparameters.
See Table~\ref{tab:Word2Vec:architecture} for an overview of the Word2Vec architecture, and Table~\ref{tab:Word2Vec:hyperparameters} for a summary of the other hyperparameters used.

\begin{table}[htbp]
\centering
\caption{The architecture of the Word2Vec-CNN fine-tuned model.}
\label{tab:Word2Vec:architecture}
\begin{tabular}{|c|c|}
\hline
\textbf{Layers}    & \textbf{Hyperparameters}                                                                                                                    \\ \hline
Embedding model    & \begin{tabular}[c]{@{}c@{}}Embedding   dimension = 300\\ Vocabulary  size = 700,000 \\ Max sentence length = 100\end{tabular} \\ \hline
Convolutional 1D             & 512, 3, activation='relu'                                                                                                                   \\ \hline
Batch Normalization &                                         N/A                                                                                                    \\ \hline
Max Pooling 1D       & 3                                                                                                                                           \\ \hline
Global Max Pooling 1D &                                         N/A                                                                                                    \\ \hline
Dense              & 512,   activation='relu'                                                                                                                    \\ \hline
Dropout            & 0.7                                                                                                                                         \\ \hline
Dense              & 6,   activation='Softmax'                                                                                                                   \\ \hline
\end{tabular}
\end{table}

\begin{table}[htbp]
\centering
\caption{The hyperparameters for the Word2Vec-CNN fine-tuned model.}
\label{tab:Word2Vec:hyperparameters}
\begin{tabular}{|c|c|}
\hline
\textbf{Hyperparameters} & \textbf{Values}                    \\ \hline
Optimizer                & Adam                               \\ \hline
Loss Function            & Sparse   categorical cross entropy \\ \hline
Learning Rate            & 1e-04                              \\ \hline
Batch size               & 256                                \\ \hline
epochs                   & 25                                 \\ \hline
\end{tabular}
\end{table}

\subsection*{GloVe}
GloVe is an unsupervised learning method that is also used to obtain vector representations of words, but with a different training process compared to Word2Vec. The training targets a word-word co-occurrence matrix, and is carried out by finding aggregated global word-word co-occurrence statistics in a corpus to capture the frequency with which words co-occur with one another~\cite{14}. 

In our experiment, a GloVe embedding model was trained on the Wikipedia 2014 + Gigaword 5 datasets (6 billion tokens, 400,000 vocab, uncased, 300 dimension vectors), and was fine-tuned using our video titles dataset. This fine-tuned GloVe model consists of 50,000 vocabularies with a maximum sentence length of 100. Each word is represented by 300 dimensions. A sequence of two Bidirectional LSTM layers was added before the dense layers. The last dense layer produced six probabilities corresponding to the six political leaning categories. This architecture is the one that gave the highest validation accuracy while tuning the hyperparameters. Table~\ref{tab:GloVe:architecture} summarizes the Glove model's architecture, while Table~\ref{tab:GloVe:hyperparameters} specifies the other hyperparameters used.

\begin{table}[htbp]
\centering
\caption{The architecture of the GloVe-LSTM fine-tuned model.}
\label{tab:GloVe:architecture}
\begin{tabular}{|c|c|}
\hline
\textbf{Layers}      & \textbf{Hyperparameters}                                                                                                                   \\ \hline
Embedding model      & \begin{tabular}[c]{@{}c@{}}Embedding   dimension = 300\\ Vocabulary   size = 50,000\\ Max   sentence length = 100\end{tabular} \\ \hline
Bidirectional   LSTM & 64                                                                                                                                         \\ \hline
Bidirectional   LSTM & 64                                                                                                                                         \\ \hline
Dense                & 6,   activation='Softmax'                                                                                                                  \\ \hline
\end{tabular}
\end{table}

\begin{table}[htbp]
\centering
\caption{The hyperparameters for the GloVe-LSTM fine-tuned model.}
\label{tab:GloVe:hyperparameters}
\begin{tabular}{|c|c|}
\hline
\textbf{Hyperparameters} & \textbf{Values}                    \\ \hline
Optimizer                & Adam                               \\ \hline
Loss Function            & Sparse   categorical cross entropy \\ \hline
Learning Rate            & 0.001                              \\ \hline
Batch size               & 256                                \\ \hline
epochs                   & 8                                  \\ \hline
\end{tabular}
\end{table}

\subsection*{BERT}

We proposed to use the state-of-the-art text classifier BERT (Bidirectional Encoder Representations from Transformers), which is based on the transformer architecture. BERT provides a dense vector representation for natural language using a deep, pre-trained neural network~\cite{15}. To train BERT, the developers used both masked language model (MLM) pre-training as well as next sentence prediction techniques. The design of BERT is based on pre-training deep bidirectional representations from unlabeled text by jointly conditioning on both right and left context in all layers. The advantage of using BERT here is the fact that the preprocessing stage is not required, given that the WordPiece tokenization technique is already involved. This technique was designed to tokenize sentences based on out-of-vocabulary words.

In our experiments, we used the BERT pre-trained preprocessor and encoder that were trained on the Wikipedia and BooksCorpus datasets. We fine-tuned it using our video titles dataset in end-to-end fashion (training all layers from the video title at the input to the political leaning category at the output), resulting in a model in which each word is represented by 768 dimensions. A sequence of a dense and dropout layers was added. The last dense layer produced six probabilities corresponding to the six political leaning categories. This architecture, which yielded the highest validation accuracy, is summarized in Table~\ref{tab:BERT:architecture}, and the other hyperparameters used are specified in Table~\ref{tab:BERT:hyperparameters}.

The implementation of BERT was done using the TF Hub model from the TensorFlow Models repository on GitHub~\cite{BERT}. It uses L=12 hidden layers (i.e., Transformer blocks), a hidden size of H=768, and A=12 attention heads. All parameters in the BERT were fine-tuned for video titles.

\begin{table}[htbp]
\centering
\caption{The architecture of the BERT fine-tuned model.}
\label{tab:BERT:architecture}
\begin{tabular}{|c|c|}
\hline
\textbf{Layers}   & \textbf{Hyperparameters}   \\ \hline
BERT   preprocess &                            \\ \hline
BERT   encoder    &                            \\ \hline
Dropout           & 0.3                        \\ \hline
Dense             & 512,   activation='relu'   \\ \hline
Dropout           & 0.3                        \\ \hline
Dense             & 1024.,   activation='relu' \\ \hline
Dropout           & 0.3                        \\ \hline
Dense             & 6,   activation='Softmax'  \\ \hline
\end{tabular}
\end{table}

\begin{table}[htbp]
\centering
\caption{The hyperparameters for the BERT fine-tuned model.}
\label{tab:BERT:hyperparameters}
\begin{tabular}{|c|c|}
\hline
\textbf{Hyperparameters} & \textbf{Values}                  \\ \hline
Embedding dimension      & 768                              \\ \hline
Optimizer                & Adam                             \\ \hline
Learning rate            & 1e-04                            \\ \hline
Loss Function            & Sparse categorical cross entropy \\ \hline
Batch size               & 128                              \\ \hline
epochs                   & 10                               \\ \hline
\end{tabular}
\end{table}

\section*{Experimental Results}

This section discusses the results after conducting several experiments to train and validate the aforementioned text classifiers---Word2Vec, GloVe, and Bert---using video titles as textual data. Here, we employed the categorization proposed by Hosseinmardi et al.~\cite{5} which consists of six classes: Far Left, Left, Center, Anti-Woke, Right, and Far Right. We trained the three classifiers with these six classes using the video title dataset~\cite{5,6,7}, and implemented them after carefully configuring the architectures and hyperparameters that yield the best performance. The results are evaluated and compared in terms of accuracy and F1 score, with a greater emphasis on F1 score due to imbalanced nature of our dataset. To qualify the upcoming analysis on the word representation of different embedding models, we first discuss the performance of the models used; see Table~\ref{tab:results:comparison} for a summary of the results.

For GloVe, we trained the model under three scenarios: (i) starting from random embedding weights and then fine-tuning on our data; (ii) utilizing the pre-trained embedding model without fine-tuning; and (iii) utilizing the pre-trained embedding model and fine-tuning on our data. In these three scenarios, we tuned the weights of the convolutional and dense layers to customize the model to fit our task and produce six political leaning categories at the output layer. As can be seen in Table~\ref{tab:results:comparison}, training from random weights (scenario~i) and using a pre-trained embedding model without fine-tuning (scenario~ii) are less efficient than fine-tuning the pre-trained GloVe model (scenario~iii); the latter yields the highest accuracy (70\%) and F1 score (72\%).

For Word2Vec, we trained the models under two scenarios: (i) utilizing the pre-trained embedding model without fine-tuning; and (ii) utilizing the pre-trained embedding model and fine-tuning on our data. In both scenarios, we tuned the weights of the bidirectional LSTM and dense layers to customize the model to fit our task and produce six political leaning categories at the output layer. As shown in Table~\ref{tab:results:comparison}, utilizing the pre-trained Word2Vec without fine-tuning is less efficient compared to fine-tuning the pre-trained Word2Vec, which has the highest accuracy (71\%) and F1 score (73\%).

Given that the fine-tuning of a pre-trained model yielded the highest accuracy and F1 score for both GloVe and Word2Vec, we followed a similar approach for BERT. This model includes a pre-processor and an encoder, both of which were fine-tuned on our dataset. Additionally, we tuned the weights of the dense layers to customize the model to fit our task and produce six political leaning categories at the output layer. As can be seen in Table~\ref{tab:results:comparison}, the fine-tuned BERT model yielded the highest accuracy (75\%) and F1 score (77\%), outperforming the other classifiers used in our experiments. This can be attributed to BERT's attention mechanism, which plays a significant role in learning powerful word and text representations. 

It is worth noting that, with every additional 1\% of accuracy, the classifier is able to correctly predict an additional 20,000 videos. As such, the fact that the fine-tuned BERT classifier achieves a 4\% increase in accuracy compared to the second-best alternative (i.e., the fine-tuned, pre-trained Web2Vec) translates to a substantial improvement in performance, as it implies that the former classifier can correctly predict an additional 80,000 videos compared to the latter. Motivated by these results, we focus on our fine-tuned BERT classifier for the remainder of this study. 

\begin{table}[htbp]
\centering
\caption{A comparison between the proposed video title classifier (BERT) and the other baseline classifiers (Word2Vec and GloVe) in terms of weighted average accuracy, precision, recall, and F1 score.}
\label{tab:results:comparison}
\begin{tabular}{|l|c|c|c|c|} \hline
\textbf{Methods} &
  \textbf{\begin{tabular}[c]{@{}c@{}}Average\\ accuracy\end{tabular}} &
  \textbf{\begin{tabular}[c]{@{}c@{}}Average \\ precision\end{tabular}} &
  \textbf{\begin{tabular}[c]{@{}c@{}}Average\\ recall\end{tabular}} &
  \textbf{\begin{tabular}[c]{@{}c@{}}Average\\ F1 score\end{tabular}} \\ \hline
GloVe trained from random & \multirow{2}{*}{0.67} & \multirow{2}{*}{0.75} & \multirow{2}{*}{0.67}          & \multirow{2}{*}{0.70}          \\ 
embedding weights (baseline)~\cite{14}  & & & & \\ \hline
Pre-trained GloVe without & \multirow{2}{*}{0.66}   & \multirow{2}{*}{0.74} & \multirow{2}{*}{0.66}    & \multirow{2}{*}{0.68} \\ 
fine-tuning (baseline)~\cite{14} & & & & \\ \hline
Pre-trained GloVe with & \multirow{2}{*}{0.70} & \multirow{2}{*}{0.77} & \multirow{2}{*}{0.70} & \multirow{2}{*}{0.72} \\ 
fine-tuning (baseline)~\cite{14} & & & & \\ \hline
Pre-trained Web2Vec without & \multirow{2}{*}{0.63} & \multirow{2}{*}{0.73} & \multirow{2}{*}{0.63} & \multirow{2}{*}{0.66} \\ 
fine-tuning (baseline)~\cite{13} & & & & \\ \hline
Pre-trained Web2Vec & \multirow{2}{*}{0.71} & \multirow{2}{*}{0.78} & \multirow{2}{*}{0.71} & \multirow{2}{*}{0.73} \\ 
with fine-tuning (baseline)~\cite{13} & & & & \\ \hline
\textbf{BERT-based classifier}                      & \multirow{2}{*}{\textbf{0.75}} & \multirow{2}{*}{\textbf{0.80}} & \multirow{2}{*}{\textbf{0.75}} & \multirow{2}{*}{\textbf{0.77}}
\\ 
\textbf{(our proposed model)} & & & & \\ \hline
\end{tabular}
\end{table}

\begin{figure}[htbp]
    \centering
    \includegraphics[width=0.7\linewidth]{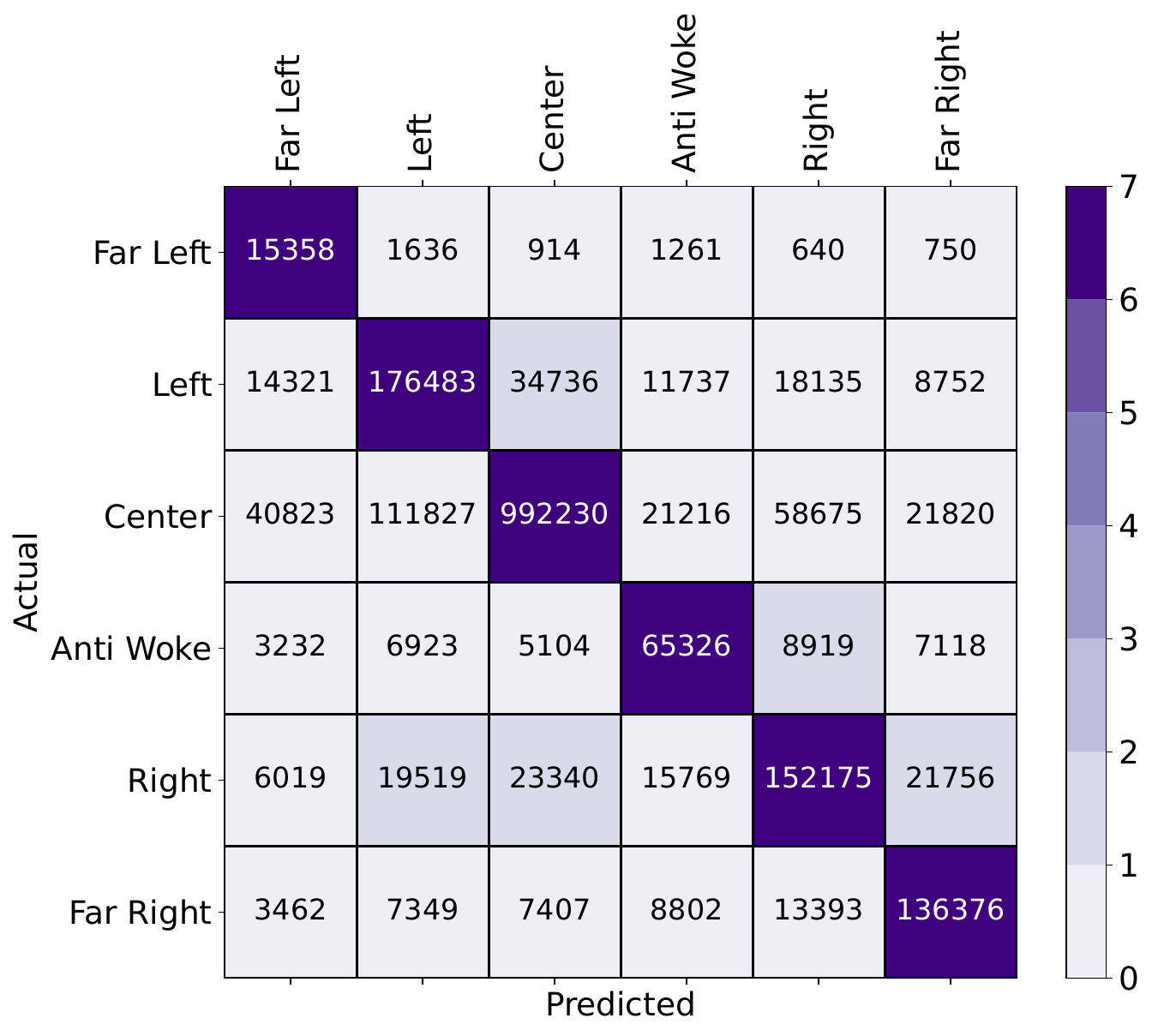}
    \caption{Confusion Matrix of predictions made by our YouTube Political Leaning Classifier.}
    \label{fig:confusion_matrix}
\end{figure}

Continuing with BERT's performance analysis, we examine its confusion matrix; see Figure~\ref{fig:confusion_matrix}. Given the imbalanced nature of our dataset, the visualization of each row is improved by splitting the range of values therein into equal bins, and assigning a different color to each bin (greater values correspond to darker colors). Looking at the confusion matrix, it becomes clear that the data is imbalanced, as the majority of samples belong to the Center category. This implies that the false predictions come largely from incorrectly classifying the videos as Center. The confusion matrix also shows that the incorrect predictions are mostly concentrated around the correct class. For example, looking at Far Right videos (bottom row), we can see that most of the incorrect predictions are actually classified as Right. While this is an incorrect classification, it is closer to the ground truth than incorrectly classifying the videos as Left or Far Left. Overall, the classifier rarely classifies right-leaning videos as left-leaning, or vice versa.

The classification report is provided in Table~\ref{tab:classification_report}, specifying the accuracy, precision, recall, and F1 score of the BERT model for each of the six political leaning categories. As can be seen, Center has the best accuracy (80\%), recall (80\%), precision (93\%), and F1 score (86\%); this is probably due to the fact that Center has the largest number of samples compared to the other categories. The second-best prediction is for Far Right; while the accuracy, recall, precision, and F1 score are all lower than the corresponding values for Center, they are all higher than the corresponding values for any of the remaining categories. The worst F1 score is for the Far Left category, probably due to the fact that it has fewer samples compared to any other category.

\begin{table}[htbp]
\centering
\caption{classification report specifying the accuracy, precision, recall, and F1 score of the BERT model for each category. The bottom row shows the weighted average, taken over all categories.}
\label{tab:classification_report}
\begin{tabular}{|c|c|c|c|c|}
\hline
\textbf{Category} & \textbf{Accuracy} & \textbf{Precision} & \textbf{Recall} & \textbf{F1 score} \\ \hline
Far Left          &0.74                   & 0.18               & 0.75            & 0.30              \\ \hline
Left              &0.67                   & 0.55               & 0.67            & 0.60              \\ \hline
Center            &0.80                   & 0.93               & 0.80            & 0.86              \\ \hline
Anti-Woke         &0.68                   & 0.53               & 0.68            & 0.59              \\ \hline
Right             &0.64                   & 0.60               & 0.64            & 0.62              \\ \hline
Far Right         & 0.77                  & 0.69               & 0.77            & 0.73              \\ \hline
Weighted average  & 0.75              & 0.80               & 0.75            & 0.77              \\ \hline
\end{tabular}
\end{table}

Having evaluated the classifiers using the testing data with 2 million video titles, we now validate the BERT model on a real-world application. In particular, given the YouTube channels of news agencies, our goal is to predict the distribution of the political leaning of the videos in each of these news channel. For ground truth labels, we utilized the Allsides media bias chart~\cite{16}, which specifies the political leaning of prominent news agencies. We selected five news agencies that are Right, five that are Center, and five that are Left, resulting in a total of 15 agencies. We collected videos from the YouTube channel of each news agency. To this end, we used the \textit{YouTube Search Python} package, which caps the number of videos per channel at around 20,000. For channels containing fewer than 20,000 videos, we collected all the videos therein. Table~\ref{tab:ground_truth} specifies the ground truth political leaning of each news agency, along with the number of videos that we collected from the YouTube channel of each agency.

\begin{table}[htbp]
\centering
\caption{The ground truth label of each news agency, and the number of videos collected from the YouTube channel of each agency.}
\label{tab:ground_truth}
\begin{tabular}{|c|c|c|}
\hline
\textbf{Ground Truth   Category} & \textbf{YouTube   channel} & \textbf{Number of   Videos} \\ \hline
\multirow{5}{*}{Center}          & Forbes                     & 6390                        \\  
                                 & The Hill                    & 19988                       \\  
                                 & Reuters                    & 19796                       \\  
                                 & The Wall Street Journal  & 19674                       \\  
                                 & BBC news                   & 19547                       \\ \hline
\multirow{5}{*}{Left}            & MSNBC                      & 19947                       \\  
                                 & CNN                        & 19268                       \\  
                                 & New York Times           & 10116                       \\  
                                 & NBC news                   & 19215                       \\  
                                 & The Guardian                   & 7126                        \\ \hline
\multirow{5}{*}{Right}           & Fox news                    & 19942                       \\  
                                 & New York post              & 12839                       \\  
                                 & CBN news                   & 19754                       \\  
                                 & Blaze media                & 11394                       \\  
                                 & News Max                    & 19778                       \\ \hline
\end{tabular}
\end{table}

\begin{figure}[htbp]
    \captionsetup[subfigure]{labelformat=empty}
    \centering    
    \begin{subfigure}{0.32\textwidth}
        \caption{CNN}
        \label{fig:cnn}
        \includegraphics[width=1\linewidth]{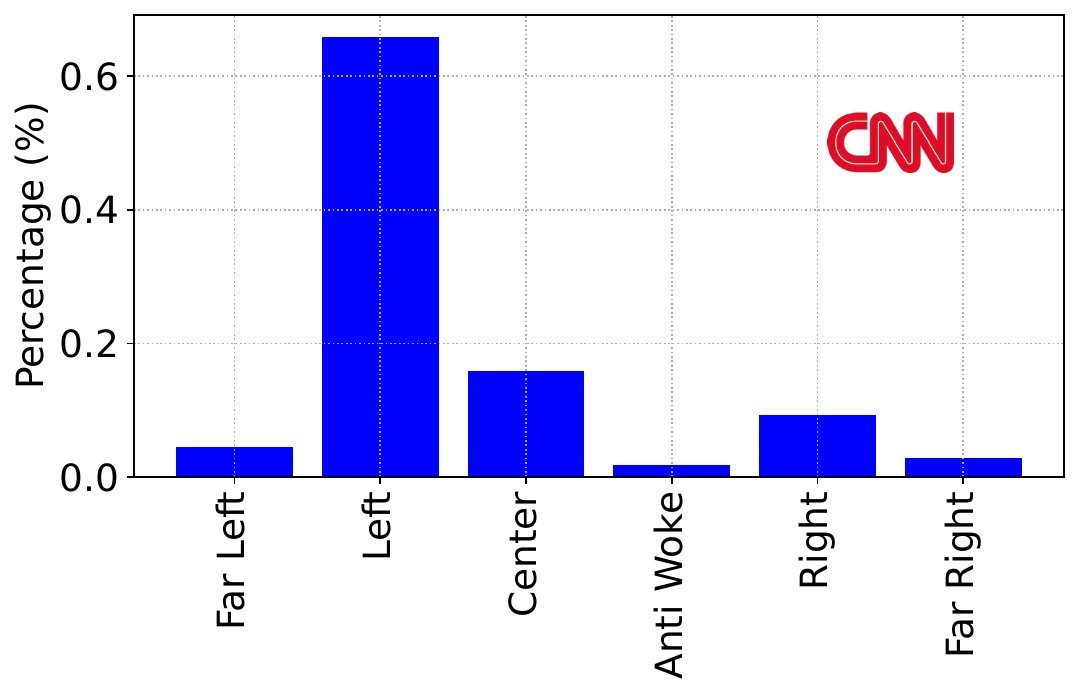}
    \end{subfigure}~
    \begin{subfigure}{0.32\textwidth}
        \caption{Reuters}
        \label{fig:reuters}
        \includegraphics[width=1\linewidth]{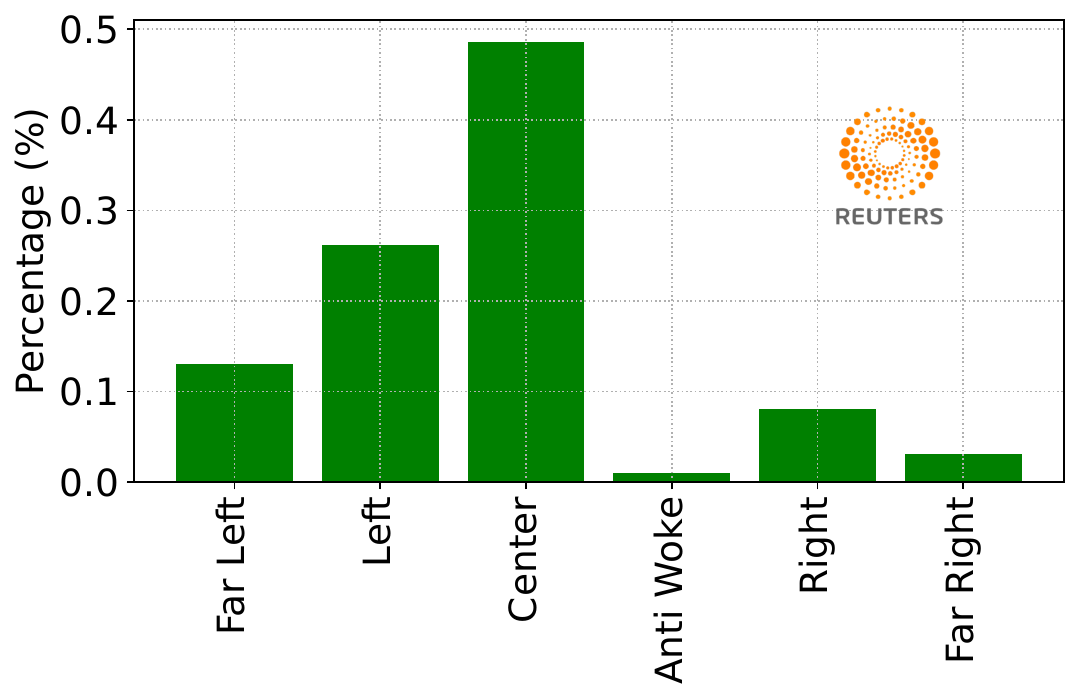}
        
    \end{subfigure}~
    \begin{subfigure}{0.32\textwidth}
        \caption{Blaze Media}
        \label{fig:blazemedia}
        \includegraphics[width=1\linewidth]{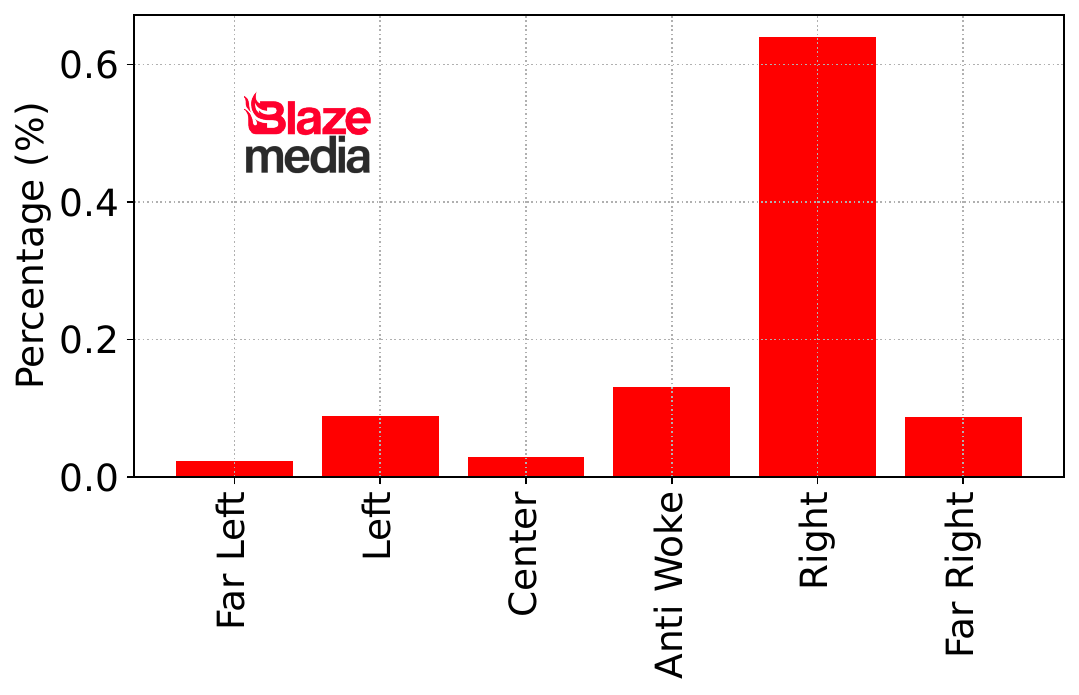}
        
    \end{subfigure}\\
    \begin{subfigure}{0.32\textwidth}
        \caption{MSNBC}
        \label{fig:msnbc}
        \includegraphics[width=1\linewidth]{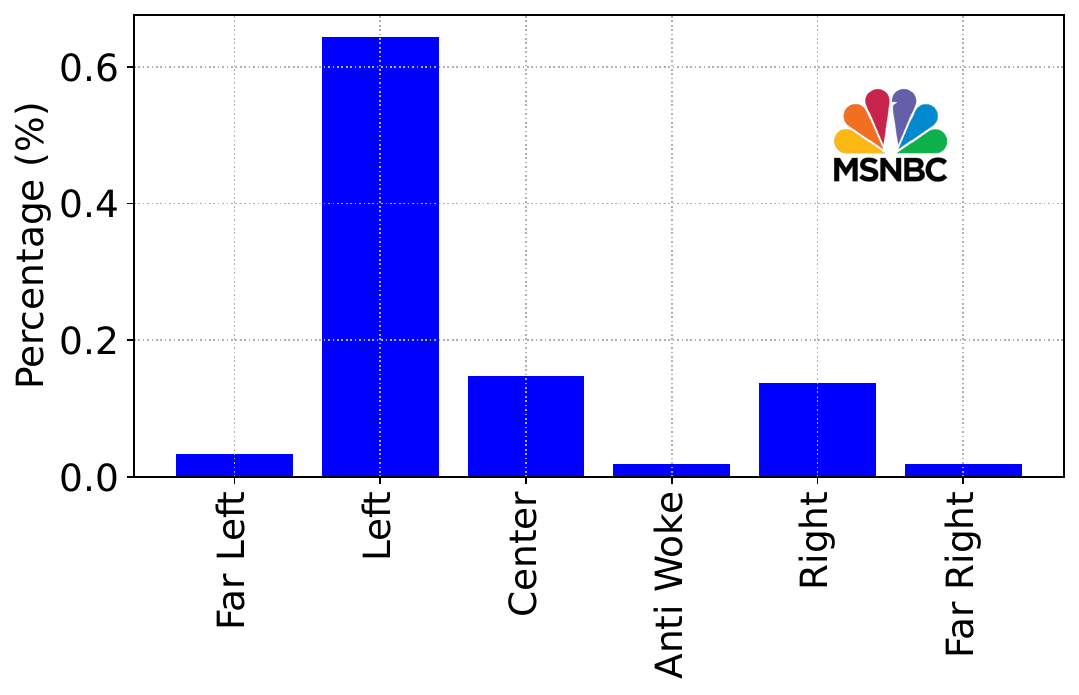}
        
    \end{subfigure}~
    \begin{subfigure}{0.32\textwidth}
        \caption{BBC News}
        \label{fig:bbcnews}
        \includegraphics[width=1\linewidth]{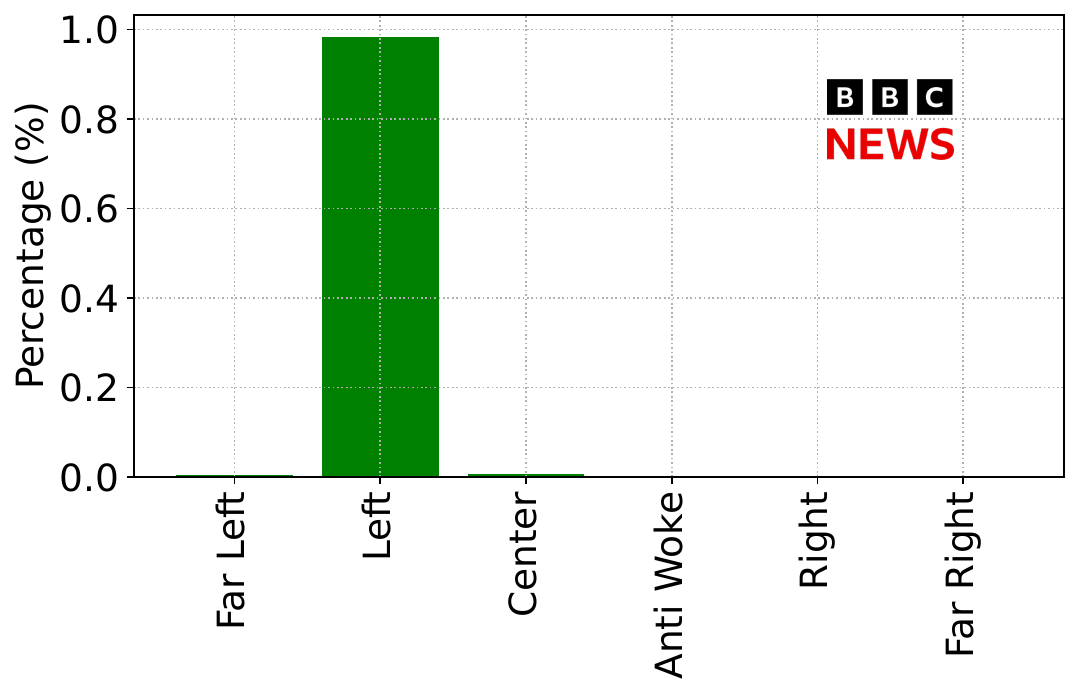}
        
    \end{subfigure}~
    \begin{subfigure}{0.32\textwidth}
        \caption{Fox News}
        \label{fig:foxnews}
        \includegraphics[width=1\linewidth]{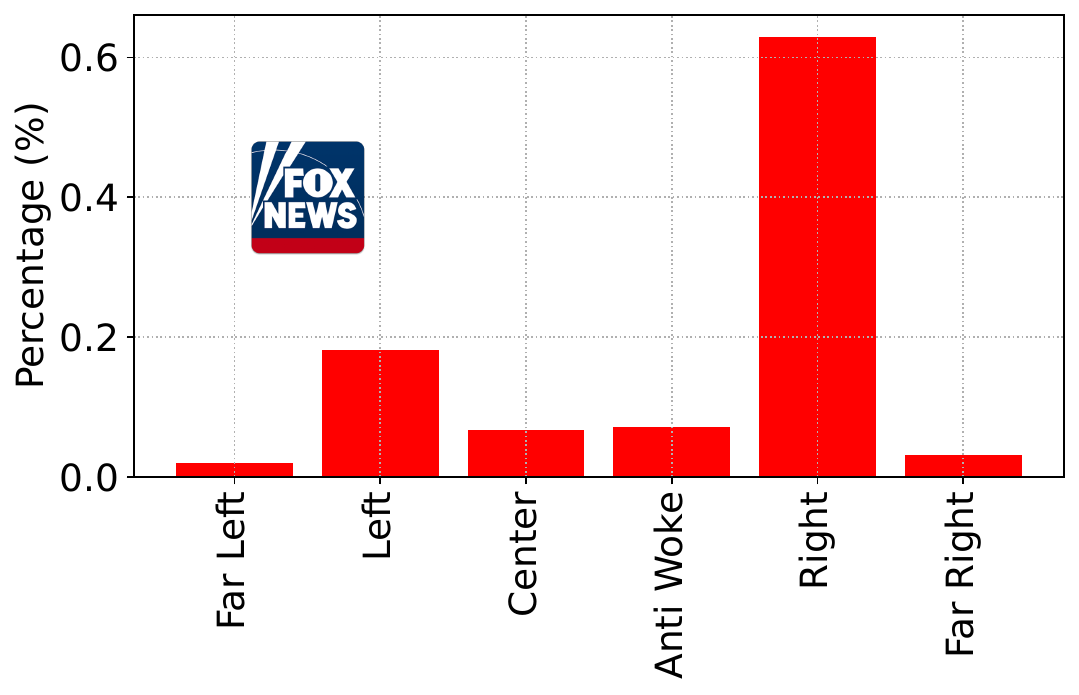}
        
    \end{subfigure}\\
    \begin{subfigure}{0.32\textwidth}
        \caption{The Guardian}
        \label{fig:guardian}
        \includegraphics[width=1\linewidth]{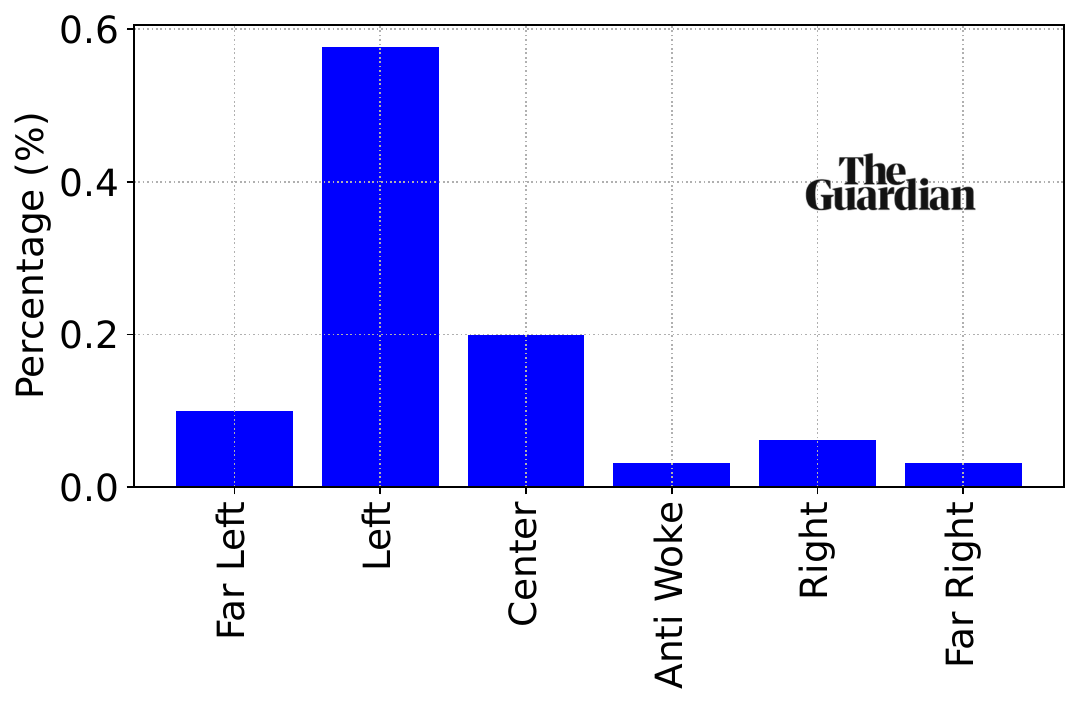}
        
    \end{subfigure}~
    \begin{subfigure}{0.32\textwidth}
        \caption{Forbes}
        \label{fig:forbes}
        \includegraphics[width=1\linewidth]{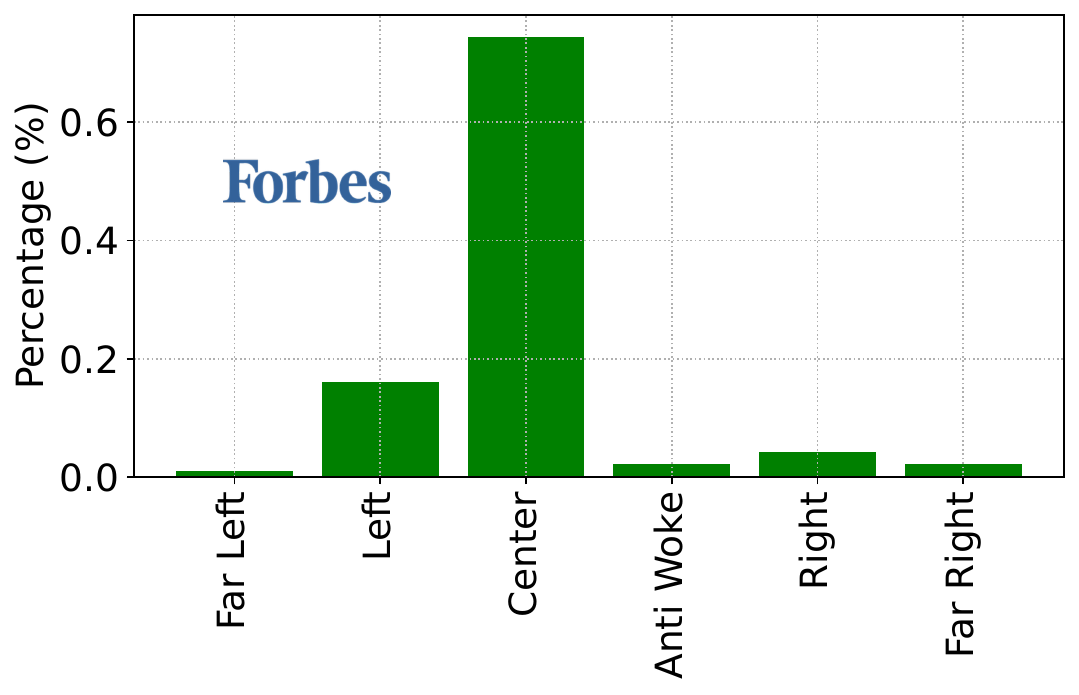}
        
    \end{subfigure}~
    \begin{subfigure}{0.32\textwidth}
        \caption{New York Post}
        \label{fig:nypost}
        \includegraphics[width=1\linewidth]{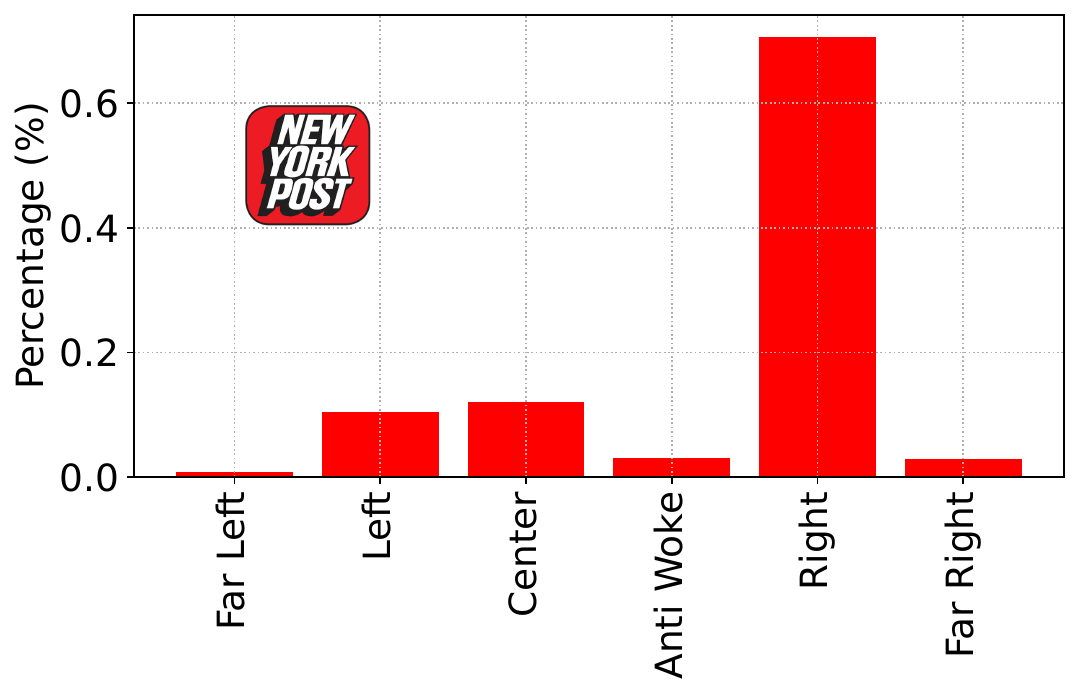}
        
    \end{subfigure}\\
    \begin{subfigure}{0.32\textwidth}
        \caption{New York Times}
        \label{fig:nytimes}
        \includegraphics[width=1\linewidth]{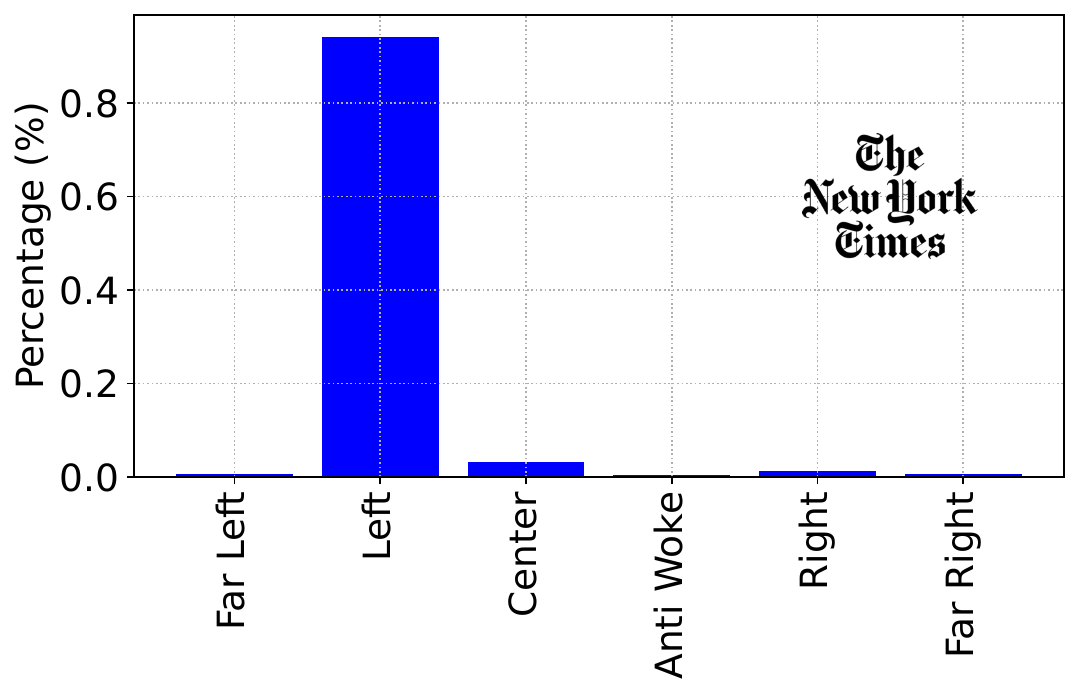}
        
    \end{subfigure}~
    \begin{subfigure}{0.32\textwidth}
        \caption{The Hill}
        \label{fig:thehill}
        \includegraphics[width=1\linewidth]{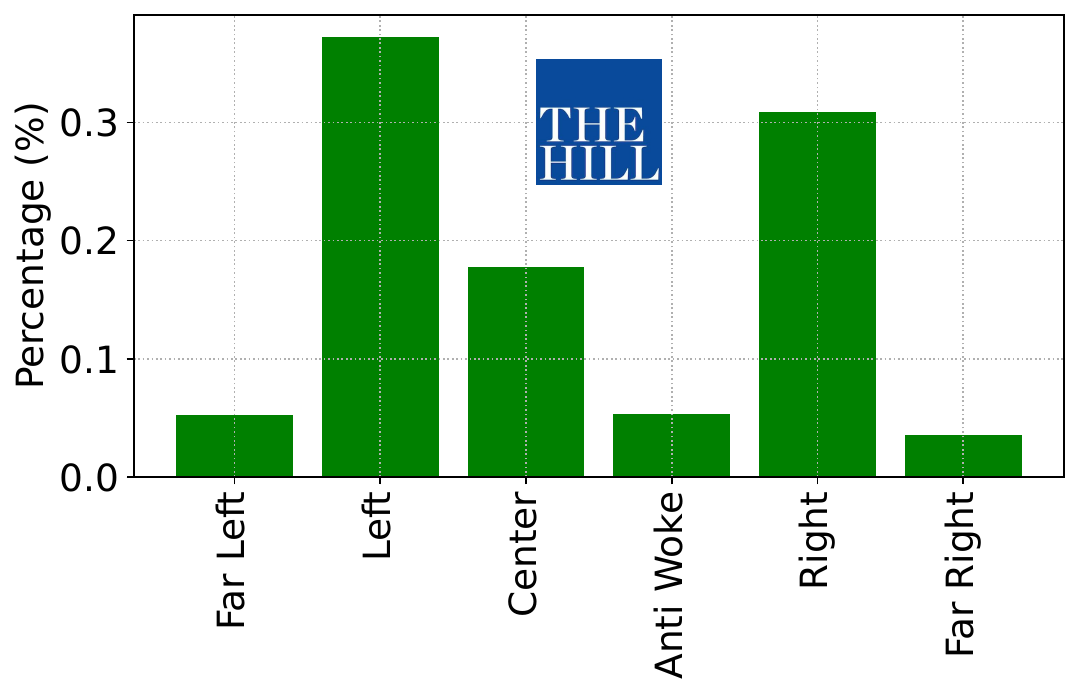}
        
    \end{subfigure}~
    \begin{subfigure}{0.32\textwidth}
        \caption{News Max}
        \label{fig:newsmax}
        \includegraphics[width=1\linewidth]{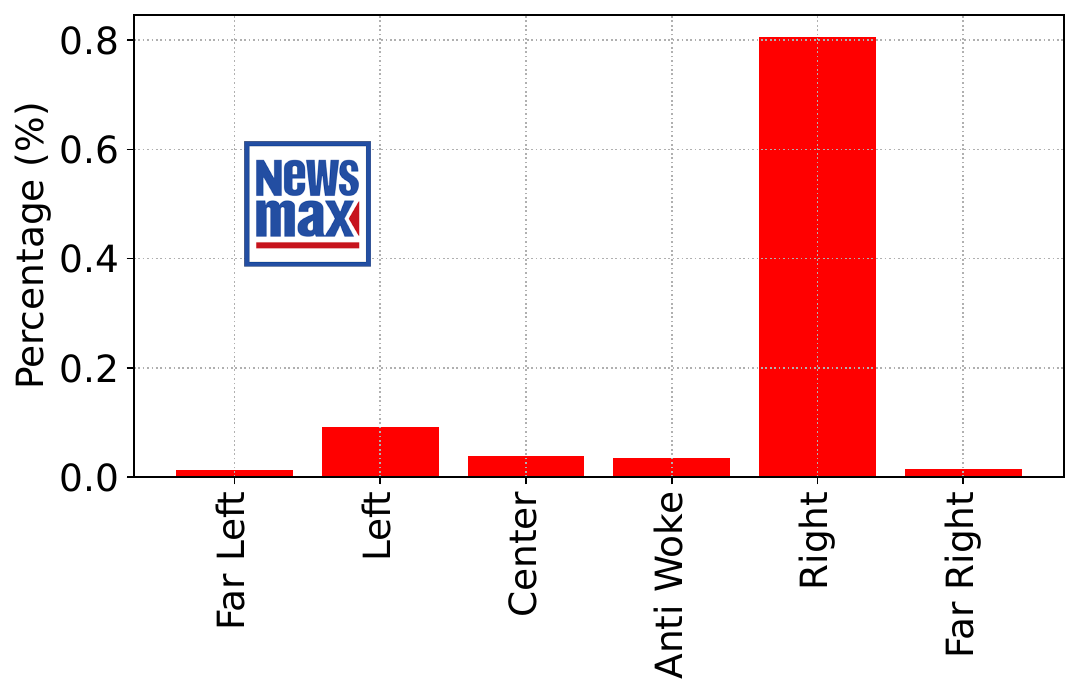}
        
    \end{subfigure}\\
    \begin{subfigure}{0.32\textwidth}
        \caption{NBC News}
        \label{fig:nbcnews}
        \includegraphics[width=1\linewidth]{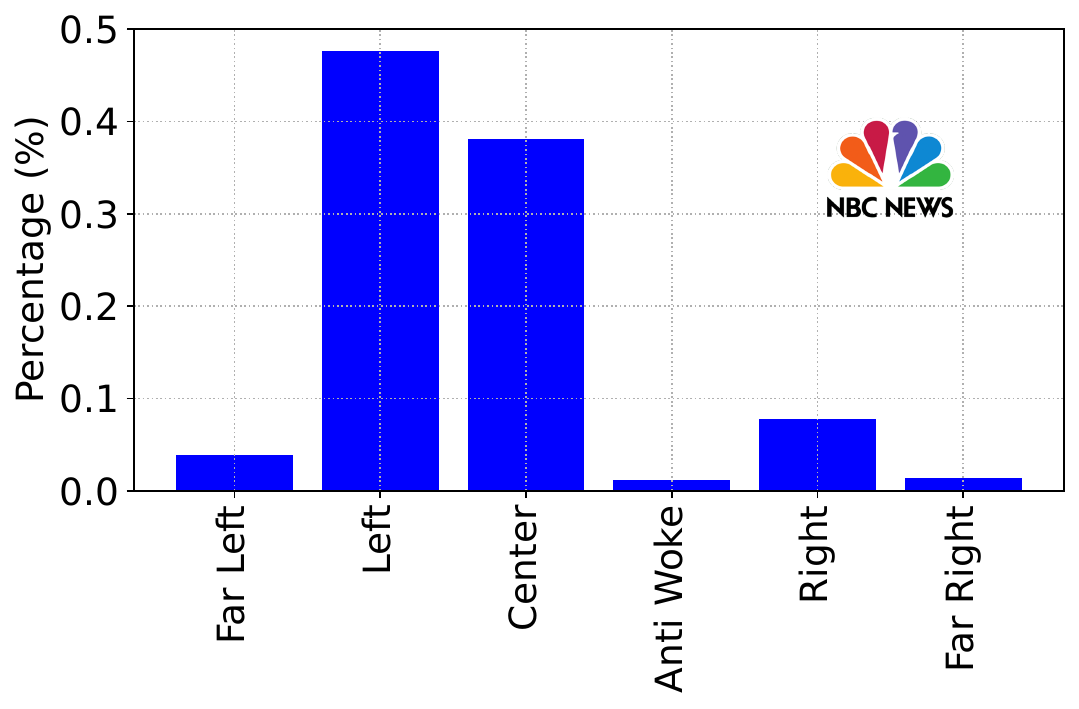}
        
    \end{subfigure}~
    \begin{subfigure}{0.32\textwidth}
        \caption{The Wall Street Journal}
        \label{fig:wallstreet}
        \includegraphics[width=1\linewidth]{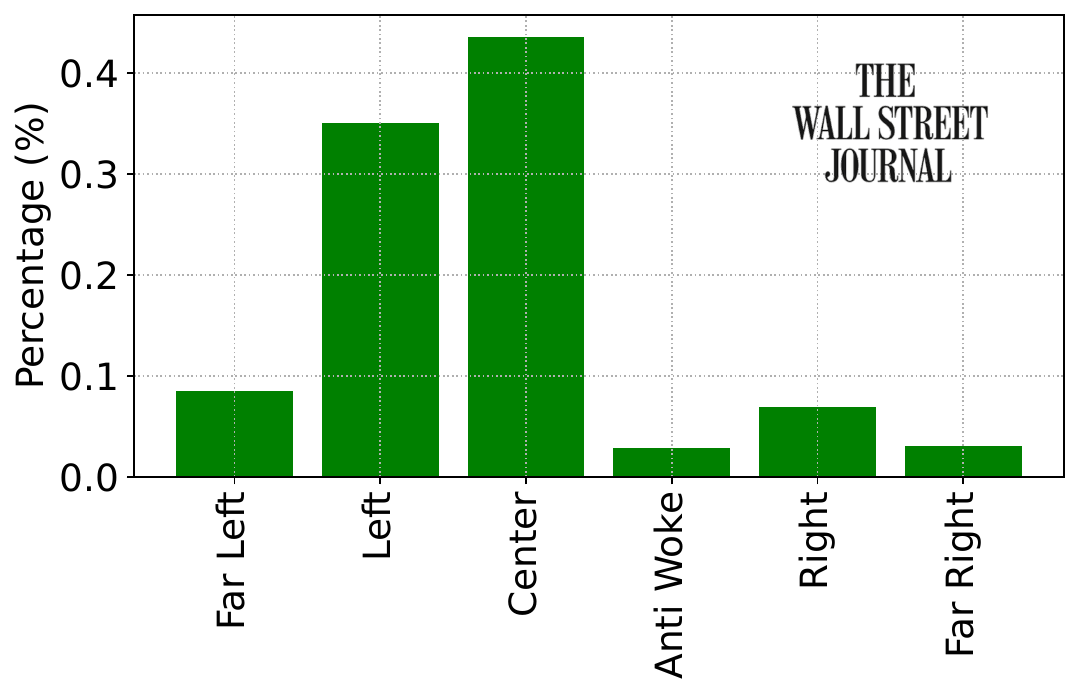}
        
    \end{subfigure}~
    \begin{subfigure}{0.32\textwidth}      
        \caption{CBN News}
        \label{fig:cbnnews}
        \includegraphics[width=1\linewidth]{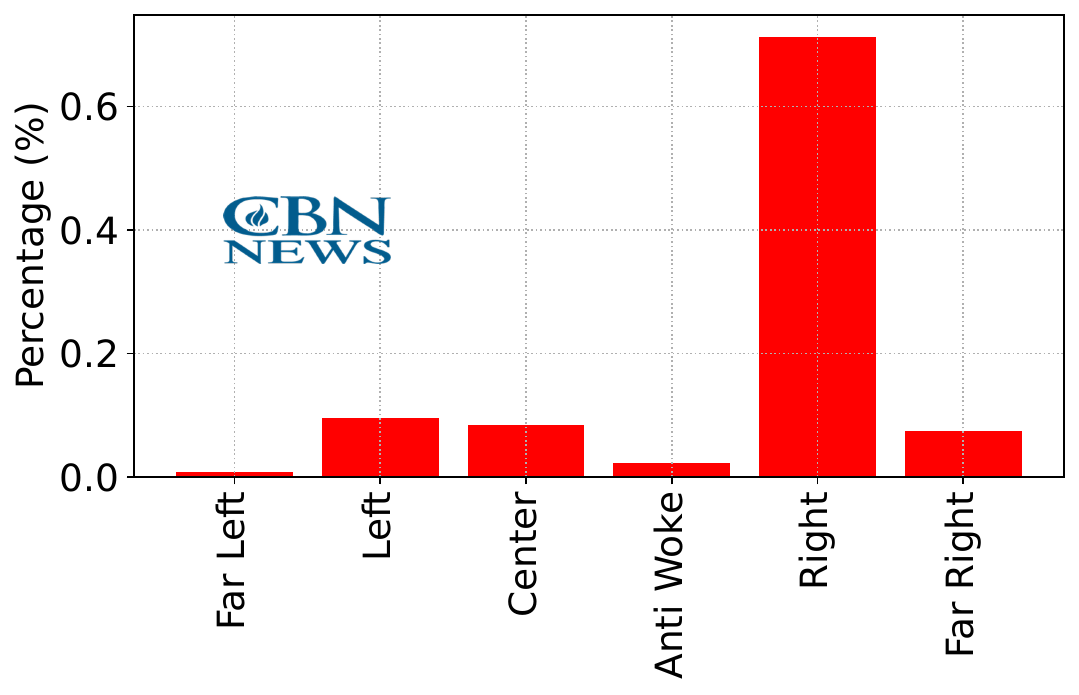}
        
    \end{subfigure}
    \caption{Distribution of political leaning predictions of videos in 15 YouTube channels. The left, center, and right columns correspond to channels whose ground truth political leaning is Left, Center, and Right, respectively.}
   \label{fig:channels}
\end{figure}

Figure~\ref{fig:channels} shows the distributions of the political leaning of videos in each of the 15 YouTube channels. As can be seen, the distribution is consistent with the ground truth political leaning for all five Left channels, as well as all five Right channels; see how the most frequent prediction in the left column is Left, and the most frequent prediction in the right column is Right. Notice that the channels from each side rarely cover content from the opposite side. However, Left channels are more likely to cover Center content than Right channels, suggesting that the former ones are less extreme. As for Center channels, the distribution is clearly consistent with the ground truth in three cases (Reuters, Forbes, and The Wall Street Journal), as the most frequent prediction for these channels is Center. As for The Hill, it can be argued that the distribution is also consistent with the ground truth. After all, if the majority of the videos in that channel are split somewhat equally between Right and Left, then the most plausible conclusion would be that the channel is neither Right nor Left, but rather covers both perspectives, thereby arguably serving as a Center channel. The only channel for which the distribution is inconsistent with the ground truth is BBC. While the channel is classified as Center according to the AllSides media bias chart, almost all its videos all classified as Left according to our classifier. 

Similarly, the wall-street-journal has leaned to Center in the last five years more than before. In blaze media, the trend towards anti-woke is seen recently associated with a decrease in right leaning. Both Blaze Media and New York Post indicate a drop in right leaning for several years. The Hill depicts both left and right leaning over years to be equal recently which may reflect its Center leaning. For several channels including CNN News, New York Times, MSNBC, The Guardian, and NBC news, the left leaning over years is clear. On the contrary, for other channels including Fox News, News Max, CBN news, the right leaning over years is obvious. 

\section*{Discussion}

This study contributes to the literature in two ways. First, we were able to utilize the transfer-learning approach by fine-tuning three pre-trained text classifiers, namely Word2Vec, Global Vectors for Word Representation (GloVe), and Bidirectional Encoder Representations from Transformers (BERT), and fine-tune them on a dataset consisting of 11.5 million video titles labelled according to their political leaning. Two million videos were reserved for testing purposes, revealing that the proposed BERT classifier has an accuracy of 75\% and F1-score of 77\%, outperforming the other baseline classifiers such as Word2Vector-CNN and GloVe-LSTM. Second, we further validated our findings by collecting thousands of videos from 15 YouTube channels of prominent news agencies with widely-known political leanings, such as Fox News and New York Times, and plotting their leaning distributions. In the vast majority of cases, the classifier outcome is consistent with the political leaning reported by the AllSides Media Bias Chart~\cite{16}. Overall, the classifier is able to detect the leaning in media, particularly videos, and classify them into six categories---Far Left, Left, Center, Anti-Woke, Right, and Far Right, using only video titles. In principle, our classifier can be a practical tool to analyze the political leaning of any YouTube channel.

There is a limitation related to the dataset used in this study. In particular, the videos in that dataset were classified solely based on their channel. However, it is not necessarily the case that all videos on a certain channel have the same political leaning as that channel. Indeed, in our evaluation of the 15 news channels, some videos had a political leaning that was different from the average leaning of the channel. In future work, one could obtain superior accuracy and prediction by training a classifier on a dataset in which every video is labelled based on its content and not just the channel it falls under.

To improve the prediction of political leaning, utilizing the transcripts of videos may be valuable, as it allows for videos with similar titles to vary in terms of their political leaning. The transcript may be overly long, and thus summarizing the transcript using AI may be required to feed the classifier with relatively short transcripts. Furthermore, we intend to extend the current study and target other video streaming platforms such as TikTok and Instagram, which are more popular among young people.


\section*{Data Availability}
All of the data used in our analysis can be found at the following repository: \url{https://github.com/comnetsAD/YouTube_political_leaning}

\section*{Author contributions statement}
N.A., T.R. and Y.Z. conceived the study, designed the research, produced the visualizations, and wrote the manuscript; N.A. performed the literature review, collected and analyzed the data, and ran the experiments.

\bibliography{sn-bibliography}

\end{document}